\documentclass[journal]{IEEEtran}

\usepackage{times}
\usepackage{epsfig}
\usepackage{graphicx}
\usepackage{amsmath}
\usepackage{amssymb}
\usepackage{overpic}
\usepackage{multirow}
\usepackage[english]{babel}

\usepackage[pagebackref=true,breaklinks=true,letterpaper=true,colorlinks,bookmarks=false]{hyperref}

\begin{document}

\title{Deep Temporal Analysis for Non-Acted Body Affect Recognition}

\author{
	Danilo~Avola,~\IEEEmembership{Member,~IEEE,}
	Luigi~Cinque,~\IEEEmembership{Senior~Member,~IEEE,}
	Alessio~Fagioli,
	Gian~Luca~Foresti,~\IEEEmembership{Senior~Member,~IEEE,}
	and~Cristiano~Massaroni,~\IEEEmembership{Student~Member,~IEEE}
	\thanks{D. Avola, L. Cinque, A. Fagioli and C. Massaroni are with the Department of Computer Science, Sapienza University, Via Salaria 113, 00198, Rome, Italy (e-mails: \{avola,cinque,fagioli,massaroni\}@di.uniroma1.it).}
	\thanks{G.L. Foresti is with the Department of Mathematics, Computer Science and Physics, University of Udine, Via delle Scienze 206, 33100, Udine, Italy (e-mail: gianluca.foresti@uniud.it).}
}
\date{}
\maketitle

\begin{abstract}
Affective computing is a field of great interest in many computer vision applications, including video surveillance, behaviour analysis, and human-robot interaction. Most of the existing literature has addressed this field by analysing different sets of face features. However, in the last decade, several studies have shown how body movements can play a key role even in emotion recognition. The majority of these experiments on the body are performed by trained actors whose aim is to simulate emotional reactions. These unnatural expressions differ from the more challenging genuine emotions, thus invalidating the obtained results. In this paper, a solution for basic non-acted emotion recognition based on 3D skeleton and Deep Neural Networks (DNNs) is provided. The proposed work introduces three majors contributions. First, unlike the current state-of-the-art in non-acted body affect recognition, where only static or global body features are considered, in this work also temporal local movements performed by subjects in each frame are examined. Second, an original set of global and time-dependent features for body movement description is provided. Third, to the best of out knowledge, this is the first attempt to use deep learning methods for non-acted body affect recognition. Due to the novelty of the topic, only the UCLIC dataset is currently considered the benchmark for comparative tests. On the latter, the proposed method outperforms all the competitors. 
\end{abstract}

\section{Introduction}
Emotions are an important aspect of our daily lives \cite{damadecartes} since they affect our cognitive processes as well as how we respond to social interactions. The affective behaviours exhibited by a person allow to convey several levels of information to an interlocutor and can be expressed via various communication channels, including language, facial expressions, or body postures. Humans are in general apt to understand and interpret the information given by these behavioural cues which are, in fact, key in human-human interaction \cite{911197}. Affective computing \cite{7435182} is the research field focused on the design of autonomous systems that try to reproduce this interpretative human ability. Over the years, this field has received great interest from the scientific community, resulting in relevant applications, such as security \cite{Greenwald89}, marketing \cite{Bagozzi1999}, or health-care \cite{Tamara2014}.

Most of the available computer vision research on affective computing deals with the emotions perceived by facial expressions \cite{4468714,Zhao2003,7374704}. This is due to the wide availability of formal models, e.g., the Facial Action Coding System (FACS) \cite{Ekman1978FacialAC}, which provides a methodical approach to treat the affect recognition task. Contrarily, the lack of formal models has strongly limited the research on emotions related to postures and body movements. Moreover, it was customary to consider the body as a simple intensity indicator of the emotions previously detected by the face \cite{Argyle88}. Thanks to later studies \cite{Meeren16518,93f290}, body analysis has been recently considered an important component in the affect recognition task and, encouraged by this new perspective, several works have started to analyse the body \cite{5740837,Kapur2005GestureBasedAC}. Despite the good results shown by these solutions, their experiments were based on acted expressions performed by actors, which tried to reproduce as genuine as possible affects. As a matter of fact, a real challenging scenario for an automatic system is provided by non-acted natural expressions \cite{5704207, 6250780}, which are more complex and less separable compared to the acted ones.

In this paper, an original deep neural architecture is proposed to classify a set of non-acted affective body movements, described as a combination of meta-features. The meta-features can be divided into two categories: a set of temporal local features that describes the movement in each time instant and a set of temporal global features that describes the movement in relation to the whole analysed time window. The proposed architecture is composed of two main branches, whose purpose is to analyse a set of meta-features. The first branch uses a Long Short Term Memory (LSTM) \cite{Hochreiter97} network to manage the time-dependent features, while the second branch uses a Multi-Layer Perceptron (MLP) \cite{Popescu_2009} network to elaborate the global features. Finally, the classification is given by a layer that merges the outputs of the two branches. Several experiments were performed on the only benchmark dataset \cite{5704207} presented in the state-of-the-art for the non-acted affect recognition task, which also reports a base rate achieved by human observers. In the network training phase, a data augmentation phase was carried out to manage those classes of emotions represented by few samples. The results on the benchmark dataset show both how our method overcomes the current state-of-the-art and how the obtained accuracy is in line with the human base rate.
\subsection{Related work}
The most accurate affect recognition systems are based on electroencephalography \cite{5349316}. Although these systems achieve excellent results, they are limited by the use of dedicated sensors that require a controlled environment. The computer vision applications, on the other hand, are more suitable to work in real and uncontrolled scenarios, thanks to the use of more versatile sensors, such as RGB, RGB-D, or thermal cameras. Most of the vision based methods involve emotion recognition by the analysis of facial expressions \cite{7374704,6130446,Mavani2017FacialER}. This is due to the presence of a great deal of labelled data in the state-of-the-art, organized in datasets, such as in \cite{8013713}. Although the face is one of the most discriminative ways to identify people's emotions, it is not always possible to capture facial expressions in large and crowded environments. This aspect motivated researchers to try other solutions, including poses and body movements.

Thanks to several studies \cite{Meeren16518,93f290}, data concerning the body, correctly labelled with emotions, are beginning to appear. In addition, the evolution of sensors and feature extraction techniques are allowing to obtain increasingly higher performance in the estimation of body poses. A valid example is reported in \cite{Kapur2005GestureBasedAC}, where several basic machine learning techniques, such as logistic regression, naive Bayes, and decision tree classifier, were used to analyse the data acquired by the VICON system. A key work was proposed in \cite{KLEINSMITH20061371}, where a Mixture Discriminant Analysis (MDA) and an unsupervised Expectation Maximization (EM) model were used to build separate cultural models for affective posture recognition. In \cite{5740837} is presented a framework to analyse affective behaviour by a small set of visual features extracted from two consumer video cameras, and an unsupervised learning method based on clustering techniques. Although all these works present good results on data recited by actors, they do not address real and un-recited scenarios. The first work that focuses on this problem is presented in \cite{5704207}, where a benchmark dataset, concerning non-acted affects extrapolated from people playing with the Nintendo Wii console, is presented. Using key body postures and an MLP network, the authors manage to get results which are close to the base rate obtained by human observers. Afterwards, the authors of \cite{6250780}, using the dataset proposed in \cite{5704207}, present an interesting collection of meta-features to train a Support Vector Machine (SVM), where these meta-features consider the body movements globally, over the entire time window. It should be noted that, about the affect recognition using the body, the majority of the state-of-the-art methods are focused only on acted data, and do not consider time-dependent features or deep learning techniques.
\subsection{Contributions}
The main contributions of the proposed work, in relation to the current literature for natural non-acted affect recognition, can be summarized in three key points:
\begin{itemize}
\item the proposal of an original combination of local and global temporal features to describe body movements both at a low level, examining a given time instant (local) as well as at a high level, considering the whole analysed time window (global);
\item the introduction of time with respect to body movement analysis as a consequence of the use of temporal local features, thus allowing a dynamic posture examination instead of a static one;
\item the introduction of deep learning, to handle body affect recognition, by leveraging a custom architecture based on merging LSTM and MLP networks to correctly manage the proposed features.	
\end{itemize}

\section{Method}
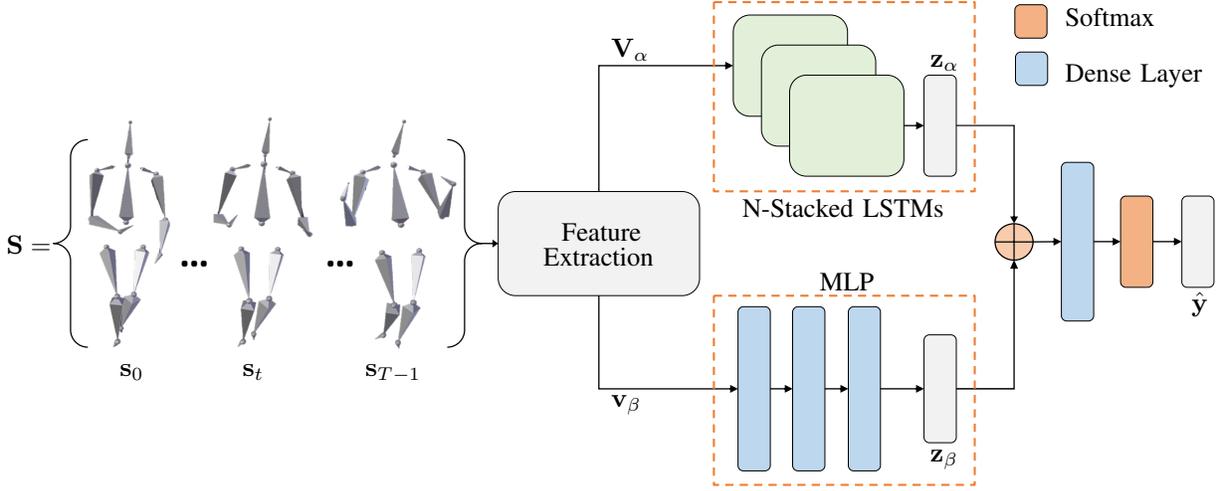
\begin{figure*}[ht]\centering
	\begin{overpic}[width=0.9\textwidth]{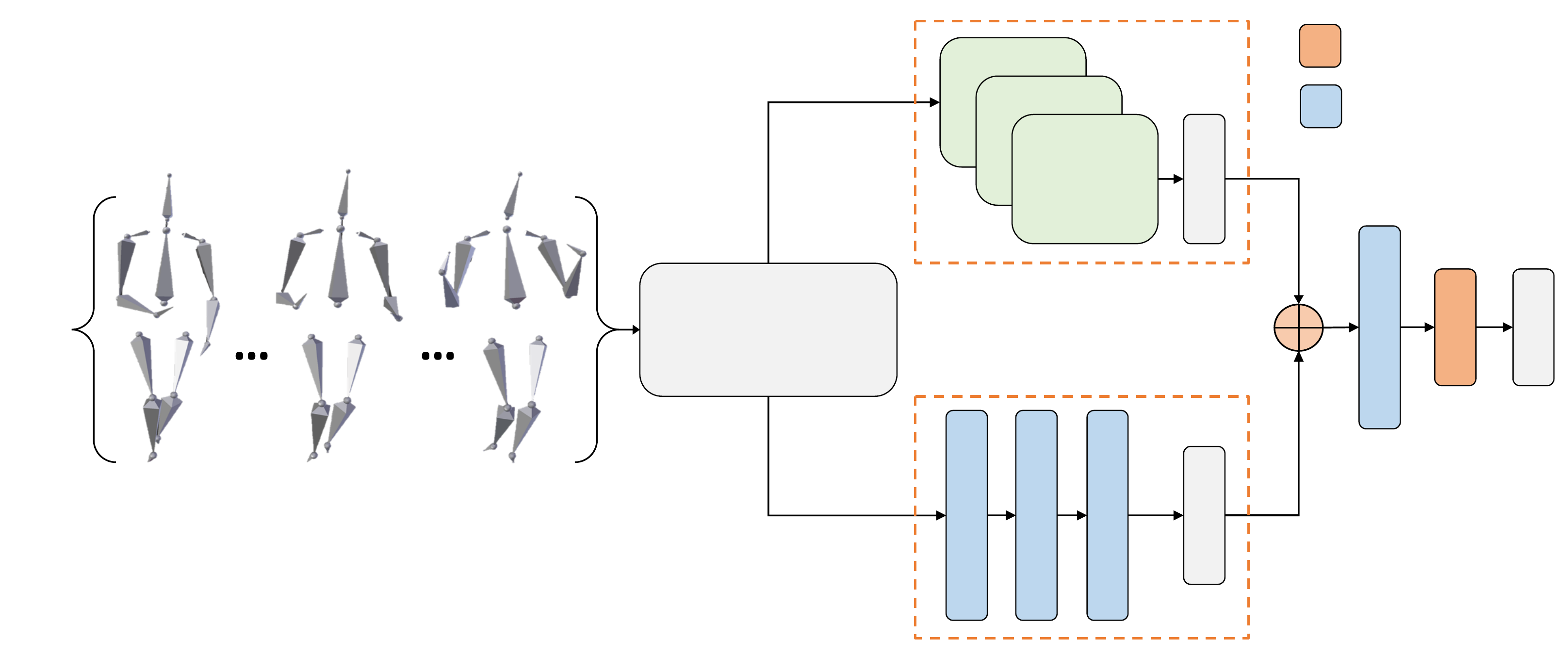}
		\put(87,38.5){Softmax}
		\put(87,34){Dense Layer}
		\put(60.7,23){N-Stacked LSTMs}
		\put(67,17){MLP}
		\put(50,36){$\mathbf{V}_\alpha$}
		\put(50,7.5){$\mathbf{v}_\beta$}
		\put(46,21){Feature}
		\put(44.5,19){Extraction}
		\put(76,35.2){$\mathbf{z}_\alpha$}
		\put(76,3){$\mathbf{z}_\beta$}
		\put(97.3,15.3){$\hat{\mathbf{y}}$}	
		\put(10,10){$\mathbf{s}_0$}
		\put(20,10){$\mathbf{s}_t$}
		\put(30,10){$\mathbf{s}_{T-1}$}
		\put(0,20){ $\mathbf{S}=$}	
	\end{overpic}
	\caption{\label{architecture} Architecture of the proposed method.}
\end{figure*}

In this section, the proposed non-acted affect recognition method based on body motions and LSTM network is described. In the first module (Sec.~\ref{feature_extraction}), the collection of temporal local feature vectors $\mathbf{V}_\alpha$ and temporal global feature vector $\mathbf{v}_\beta$ are extracted from a frame sequence $\mathbf{S}=\{\mathbf{s}_0,\dots,\mathbf{s}_t,\dots,\mathbf{s}_{T-1}\}$, representing joints rotations of a 3D skeleton for each time instant $t \in [0,T-1]$. Afterwards, $\mathbf{V}_\alpha$  and $\mathbf{v}_\beta$ are given as inputs to the two branches of the proposed network (Sec.~\ref{Proposed_Network}). Finally, the output vectors of the two branches, $\mathbf{z}_\alpha$ and $\mathbf{z}_\beta$, are combined and used to obtain the normalized vector $\hat{\mathbf{y}}$ containing the classification. In Fig.~\ref{architecture}, the overall architecture is reported. The different stages of the pipeline are detailed below.
\subsection{Feature extraction}\label{feature_extraction}
The meta-features used in this work are divided into two main categories: the temporal global features and the temporal local features.
\begin{itemize}
	\item The temporal global features are computed over the whole input sequence $\mathbf{S}$, thus describing the body movement globally without taking into account the time. Following the description of \cite{6250780}, from which we taken inspiration, this set of features can be divided into four different subsets: posture group, limb rotation movement, posture movement group, and joint rotation group. In this work, we only use the first three subsets of features. This choice derives from the fact that rotation angles along the X, Y, Z axes refer to a single key posture, as described in the joint rotation group. In our work, the rotation angles are described locally using the second feature category (i.e., temporal local features). A further difference regards the posture group, which is extended with five additional features: pose difference, pose symmetry, directed symmetry for the arm joints, as well as arms-shoulders openness.
	\item The temporal local features are computed on each 3D acquisition $\mathbf{s}_t \in \mathbf{S}$ and describe the variation of body movements in relation to time. Starting from the previously mentioned groups, these features are extracted as follows: from the first three groups (i.e., extended posture group, limb rotation movement, and posture movement group), we compute the features using two consecutive frames $\mathbf{s}_t$ and $\mathbf{s}_{t+1}$, with $\mathbf{s}_t, \mathbf{s}_{t+1} \in \mathbf{S}$ and $\forall t \in [0, T-2]$; while from the last one (i.e., joint rotation group), we take into account the normalized joint rotation angles over each single frame $\mathbf{s}_t \in \mathbf{S}$.
\end{itemize}

The final result of this feature extraction step is represented by a collection of vectors $\mathbf{V}_\alpha$ and a single vector $\mathbf{v}_\beta$, i.e., temporal local features and temporal global features, respectively.
\subsection{Proposed network architecture}\label{Proposed_Network}
The proposed network is composed of two branches: an upper branch, based on stacked LSTMs, and a lower branch, based on a MLP, operating on the features $\mathbf{V}_\alpha$ and $\mathbf{v}_\beta$, respectively. The upper branch consists of $N$-stacked LSTMs to obtain a higher-level of abstraction on the input data~\cite{Hermans}. The choice to use the LSTM derives from the remarkable results obtained in several issues regarding the analysis of body movements, such as gesture recognition \cite{8410764} or action recognition \cite{Sun_2017_ICCV}. For each time instant $t = 0,\dots,T-1$, the activation functions for an LSTM unit at the $l$-th stack layer are the following:
\begin{align}
&\resizebox{0.9\hsize}{!}{$\mathbf{i}_{l,t} = \sigma( \mathbf{W}_{xi}\mathbf{x}_{t} + \mathbf{W}_{hi}\mathbf{h}_{l,t-1} + \mathbf{W}_{ci}\mathbf{c}_{l,t-1} + \mathbf{b}_i )$}\\
&\resizebox{0.9\hsize}{!}{$\mathbf{f}_{l,t} = \sigma( \mathbf{W}_{xf}\mathbf{x}_{t} + \mathbf{W}_{hf}\mathbf{h}_{l,t-1} + \mathbf{W}_{cf}\mathbf{c}_{l,t-1} + \mathbf{b}_f )$}\\
&\resizebox{0.9\hsize}{!}{$\mathbf{c}_{l,t} = \mathbf{f}_t\odot \mathbf{c}_{t-1} + \mathbf{i}_{l,t}\odot \tanh(\mathbf{W}_{xc}\mathbf{x}_{t}+\mathbf{W}_{hc}\mathbf{h}_{l,t-1}+\mathbf{b}_c)$}\\
&\resizebox{0.9\hsize}{!}{$\mathbf{o}_{l,t} = \sigma( \mathbf{W}_{xo}\mathbf{x}_{t} + \mathbf{W}_{ho}\mathbf{h}_{l,t-1} + \mathbf{W}_{co}\mathbf{c}_{l,t-1} + \mathbf{b}_o )$}\\
&\resizebox{0.4\hsize}{!}{$\mathbf{h}_{l,t} = \mathbf{o}_{l,t} \odot \tanh(\mathbf{c}_{l,t})$}
\end{align}
where $\mathbf{i}, \mathbf{f}, \mathbf{c}$, and $\mathbf{o}$ denote input gate, forget gate, output gate, and cell activation vectors, respectively. Moreover, $\mathbf{h_{\omega}}$ vectors model the hidden states (with $\mathbf{\omega \in \{i,f,o,c\}}$) and $\mathbf{W_{h_{\omega}}}$ terms indicate their linked weight matrices. Matrices $\mathbf{W}_{xi}$, $\mathbf{W}_{xf}$, $\mathbf{W}_{xo}$, and $\mathbf{W}_{xc}$ encode the weights of input gate, forget gate, output gate, and cell; while $\mathbf{W}_{ci}$, $\mathbf{W}_{cf}$, and $\mathbf{W}_{co}$ are the diagonal weights for peep-hole connections. Finally, $\mathbf{b}_i$, $\mathbf{b}_f$, $\mathbf{b}_c$, and $\mathbf{b}_o$ denote input, forget, cell, and output bias vectors, respectively. Function $\sigma$ is the logistic sigmoid and $\odot$ denotes the element-wise product.

According to the classical LSTM architecture, each unit of an $LSTM_l$, at time $t$, takes as input a vector $\mathbf{x}_t$ and the previous hidden state $\mathbf{h}_{l,t-1}$. Where the given input $\mathbf{x}_t$ indicates a temporal local feature vector $v_t \in V_\alpha$ if $l=0$ (i.e., the first level of the stack), otherwise it represents the hidden vector of the underlying layer $\mathbf{x}_t = h_{l-1,t}$ (i.e., for layers higher than the first one).
The output vector $\mathbf{z}_\alpha$ of the upper branch is represented by $\mathbf{h}_{N-1,T-1}$, namely the hidden state of the last layer $N-1$ at the last time instant $T-1$ for the analysed time window.

Regarding the lower branch, it is composed by a MLP with 3 hidden layers, where each hidden node transforms its input, obtained from the weighted sum of the output values of the previous layer, with a rectified linear unit (ReLU) activation function. The first hidden layer has weighted connections with the temporal global feature vector $\mathbf{v}_\beta$, which represents the input layer. The number of hidden nodes in each layer (as well as the number of output nodes) is smaller than the number of input entries, in this way the MLP is used to extrapolate highly significant patterns from the input, mapping $\mathbf{v}_\beta$ into a low-dimensional description represented by the output layer vector $\mathbf{z}_\beta$.

In the last part of the network, the $\mathbf{z}_\alpha$ and $\mathbf{z}_\beta$ vectors are combined in a new vector called $\mathbf{z}$, using the concatenation operator. Afterwards, a dense layer using a ReLU activation function is applied, thus connecting each entry value in $\mathbf{z}$ to an entry of the output vector $\mathbf{y}$ via a weight. This layer is used to map the vector $\mathbf{z}$ to a number of output nodes equal to the size of the set of affects to be recognized. The size of this set is indicated with the value $K$. The final classification $\hat{\mathbf{y}}$ is obtained by applying to $\mathbf{y}$ a softmax regularization:
\begin{equation}
\hat{\mathbf{y}}(k) = \frac{e^{{\mathbf{y}}(k)}}{\sum^{K-1}_{q=0}e^{{\mathbf{y}}(q)}}\,.
\end{equation}

Finally, the proposed network is trained using the cross-entropy loss and the RMSprop optimization algorithm \cite{Duchi}.

\section{Experimental results}
In this section, implementation details and exhaustive experiments on several configurations of the proposed network are reported. We also compare our method with key works of the non-acted affective recognition literature.
\subsection{Dataset}
The benchmark used in the experiments is the UCLIC Affective Body Posture and Motion database \cite{5704207}. This collection is composed by sequences capturing 11 standing humans playing at Nintendo Wii Sports games for a minimum of thirty minutes. The chosen format is the BioVision Hierarchy (BVH), which provides both a skeleton hierarchy and its motion data. The hierarchy comprises all relevant joints for the skeleton, including head, neck, chest, collar, shoulders, elbows, wrists, hips, knees, and ankles. About the motion data, for each frame of the recording, the starting X, Y, and Z positions of the hierarchy root are stored, as well as the various joints rotations. Starting from these values, the 3D location of all body joints are computed.

Kleinsmith et al. \cite{5704207} analysed the whole dataset and hand-picked 103 frames representing skeleton configurations in which emotional expressiveness can be identified, thus defining key postures inside the available recordings. The authors, following an established protocol, labelled the frames of the UCLIC dataset according to human observations. The affects identified inside the sequences, observable during the playing games, were classified as: concentration, triumph, frustration, and defeat. Finally, human observations were used to define a base rate for human capabilities, thus estimating the classification accuracy for each of these affect classes.
\subsection{Implementation details}
All the experiments were performed on an 6-Core Intel i7 2.60GHz CPU with 32GB RAM with a GeForce GTX 1070 graphics card, while, the proposed network was implemented using the TensorFlow \cite{tensorflow2015-whitepaper} framework. These assessments were performed using the 10-cross validation average, following the protocol used in \cite{5704207}.

Both LSTM and MLP branches use 64 hidden units in each layer, where the LSTM consists in 3-stacked layers and the MLP is composed of 3 hidden dense layers. To avoid the over-fitting problem, a 30\% dropout \cite{Srivastava} probability is applied in each layer of both branches. In the stacked LSTM the dropout is the recurrent one. The training step was performed using 1500 epochs, with a learning rate of $0.01$ and a batch size of $5$. The choice of the LSTM units is based on the results obtained by comparing the most common recurrent units used in the state-of-the-art, as shown in Tab.~\ref{recurrent-unit-comparison}. This procedure was performed because, as reported in several works (e.g., \cite{Jozefowicz2015,Chung2014EmpiricalEO}), the performance of gated recurrent units may depend heavily on both dataset and corresponding task. So a trial test phase was required to choose the right cell. Although the more recent units (i.e., GRU~\cite{Jozefowicz2015}, UGRNN~\cite{Collins2017CapacityAT}, and NAS~\cite{ZophL16}) have obtained higher scores on single affective categories, such as concentrated and defeated, the LSTM was chosen since able to achieve the best average percentage overall.
\begin{table}[t]
	\begin{center}
		\resizebox{\columnwidth}{!}{
		\begin{tabular}{l c c c c}
			\hline
			Unit & C & T & F & D \\
			\hline
			\hline
			LSTM & 70.00\% & \textbf{62.86\%} & \textbf{20.00\%} & 60.00\% \\
			RNN & 64.45\%	& 48.57\% & 20.00\%	& \textbf{72.00\%}
			\\
			GRU & 76.67\% & 31.43\% & 10.00\% & 40.00\% \\
			UGRNN & \textbf{77.78\%} & 51.43\% & 0.00\% & 52.00\% \\
			NAS  & 72.22\% & 52.42\% & 0.00\% & 64.00\% \\
			\hline
		\end{tabular}
		}
	\end{center}
	\caption{Experimental results for each UCLIC dataset class, using the 10-cross validation protocol, obtained by varying recurrent neural network units; where C = concentrating, T = triumphant, F = frustrated, and D = defeated.}
	\label{recurrent-unit-comparison}
\end{table}
\subsection{Model and feature analysis}
A key aspect of the proposed method was the feature selection. The description of several feature groups, tested out to find the most effective combination, is reported bellow:
\begin{itemize}
	\item[M0:] this set is composed of 132 entries, representing all the temporal global features;
	\item[M1:] this set is composed of 101 vectors, representing the first three groups of the temporal local features computed from each frame inside the time window. Starting from the key pose indicated in the referenced dataset, the time window is built using the preceding and subsequent 50 frames;
	\item[R0:] this set represents the raw Euler joint rotation angles stored in the UCLIC dataset. This set of rotations uses the same time window of M1;
	\item[R1:] this set represents the R0 features normalized in $[0,1]$ taking into account the joint movement range. This set of rotations uses the same time window of M1.
\end{itemize}
\begin{table}[t]
	\begin{center}
		\resizebox{\columnwidth}{!}{
			\begin{tabular}{c c c c c c}
				\hline
				Model & Features & C & T & F & D \\
				\hline
				\hline
				B2 & M0 & 67.78\% & 62.86\% & 15.00\% & 62.00\% \\
				B1 & R0 & 70.00\% & 62.86\% & 20.00\% & 60.00\% \\
				B1 & R1 & 72,23\% & 65.71\% & 20.00\% & 64.00\% \\
				B1 & R0+M1  & 75.56\% & 65.71\% & 30.00\% & 64.00\% \\
				B1 & R1+M1 & 76.67\% & 65.71\% & 20.00\% & 64.00\% \\
				B1 + B2 & R0+M1, M0  & 77.78\% & 68.57\% & 30.00\% & 64.00\% \\
				B1 + B2 & R1+M1, M0 & \textbf{78.89\%} & \textbf{68.57\%} & \textbf{35.00\%} & \textbf{68.00\%} \\
				\hline
			\end{tabular}
		}
	\end{center}
	\caption{Experimental results for each UCLIC dataset class, obtained by using the first branch B1 (based on the stacked LSTM), the second branch B2 (based on the MLP), and the whole network, varying the features groups.}
	\label{feature-comparison}
\end{table}

\begin{figure}[ht]\centering
	\begin{overpic}[width=\columnwidth]{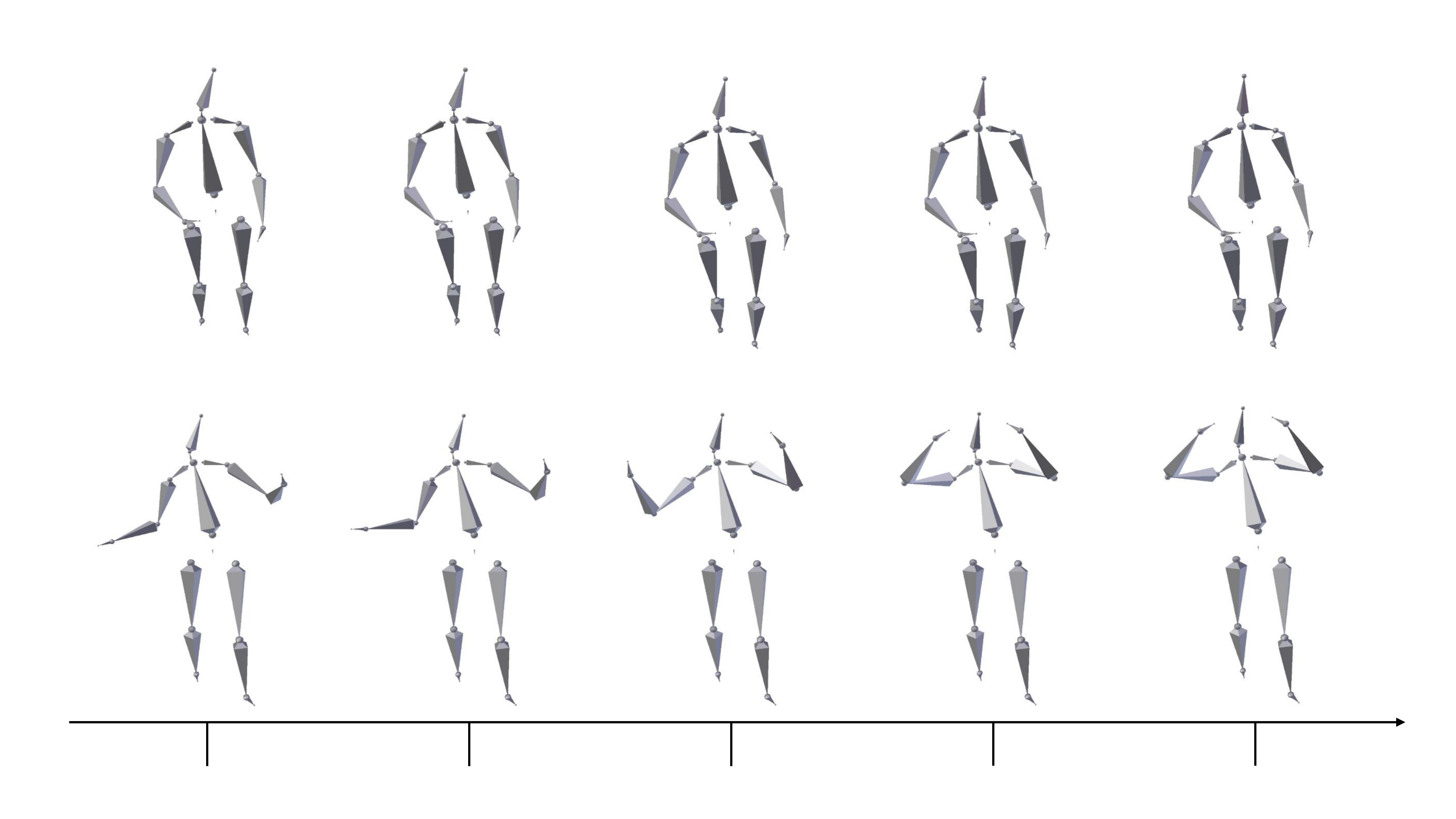}
		\put(2,41){C}
		\put(2,17){T}	
		\put(81.5,0){$t=100$}
		\put(63.5,0){$t=75$}
		\put(46.5,0){$t=50$}
		\put(29,0){$t=25$}
		\put(11,0){$t=0$}
	\end{overpic}
	\caption{\label{affect_velocity} Body motion comparison between concentrated and triumphant affects on the 101 analysed time window.}
\end{figure}

\begin{figure}[t]
	\centering
	\includegraphics[width=0.75\columnwidth]{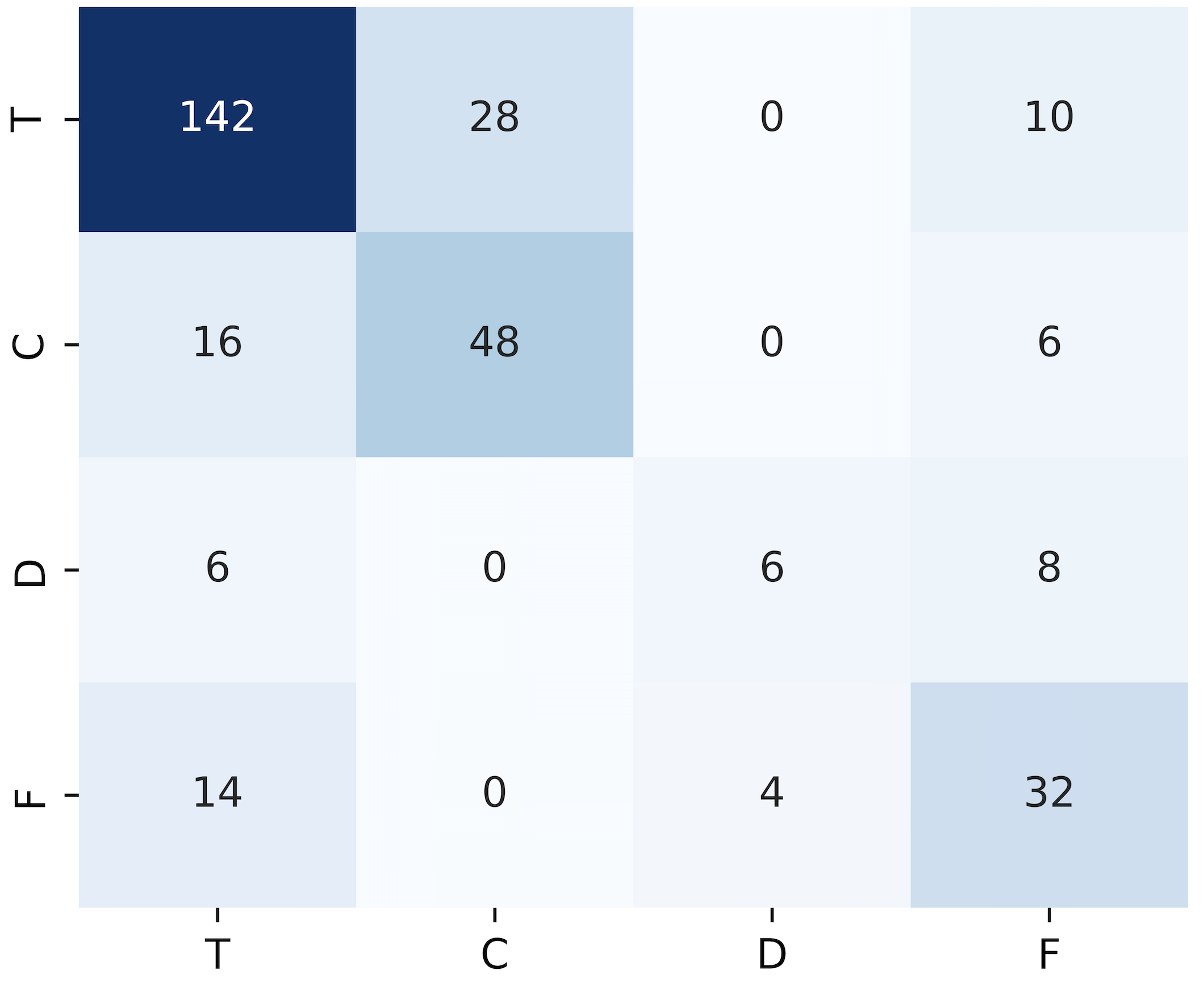}
	\caption{\label{confusion_matrix} Confusion matrix of the proposed solution according to the affect set on the UCLIC dataset over a 10-cross validation run. The rows and columns of the confusion matrix denote predictions and ground truth, respectively.}
\end{figure}

The obtained results, shown in Tab.~\ref{feature-comparison}, highlight how the simultaneous use of the two branches is more effective with respect to the single branches. Moreover, tests emphasize a better classification of the concentrated, frustrated, and defeated affects whenever the normalized rotations are considered inside the temporal local features on both single LSTM branch and full architecture. An important aspect, also backed-up by Tab.~\ref{feature-comparison}, is the relevance of time-based information. It should be noted that all the LSTM based models exceed the MLP performance, which does not consider time information. A rationale behind this behaviour can be attributed, for example, to speed variations on the human joint configuration depending on the current affective state. Indeed, as shown in Fig.~\ref{affect_velocity}, a comparison between triumphant and concentrated affects can be considered, where the former causes the human subject to perform more dynamic movements, while the latter leads to a more static motion sequence. Finally, an overview of the system classification performance, over the UCLIC dataset, is shown in Fig.~\ref{confusion_matrix}. As expected, most of the results lie on the confusion matrix diagonal. The misclassified samples are usually associated with the concentrated class, due to human observers using this affect as the neutral one in the referenced dataset. During the training step, the network learns this pattern and, following the human behaviour, uses the concentrated category as the default one.
\subsection{Data augmentation}
A common issue that afflicts numerous datasets is the low number of labelled samples. In particular, the UCLIC dataset suffers this aspect, as observable by the class related to the frustrated affect, thus making the correct classification by automated systems a hard task. To overcome this problem, an automatic data augmentation procedure was devised with the aim to increase the number of samples and to obtain a uniform distribution of data among the various emotion categories. The conceived procedure was designed as follows: starting from the original 3D skeleton joint rotation sequences of the UCLIC dataset, noise was applied to frames near the labelled key postures, following a uniform Gaussian distribution with low standard deviation (i.e., $\pm 5^{\circ}$ per axis angle), thus generating new joint rotation sequences. This new motion data was then normalized to avoid meaningless rotations, which are, normally, either unfeasible or extremely rare (e.g., an head turned backwards, or an hyper-extended elbow) to be used inside the dataset.
\begin{table}[t]
	\begin{center}
		\resizebox{\columnwidth}{!}{
			\begin{tabular}{c c c c c c}
				\hline
				Type of Dataset &Features & C & T & F & D \\
				\hline
				\hline
				
				\multirow{ 4}{*}{Base} 
				&R0& 70.00\% & 62.86\% & 20.00\% & 60.00\% \\
				&R1 & 72,23\% & 65.71\% & 20.00\% & 64.00\% \\
				&R0+M1, M0 & 77.78\% & 68.57\% & 30.00\% & 64.00\% \\
				&R1+M1, M0 & 78.89\% & 68.57\% & 35.00\% & 68.00\% \\
				\hline
				\multirow{ 4}{*}{Augmented}
				&R0 &  73.28\% & 67.59\% & 10.00\% & 61.00\% \\
				&R1 & 74.95\% & 67.82\% & 20.00\% & 61.25\% \\
				&R0+M1, M0 & 76.67\% & 79.05\% & 50.84\% & 67.00\% \\
				&R1+M1, M0 & 82.22\% &  80.95\% &  51.67\% & 70.00\% \\
				\hline
			\end{tabular}
		}
	\end{center}
	\caption{Experimental result comparison, for each UCLIC dataset class, among the various proposed network models, with or without the augmented data, using the 10-cross validation protocol.}
	\label{augemented-table}
\end{table}

With this simple approach the procedure was able to generate synthetic data sequences that are consistent with the real data, thus achieving a uniform distribution of samples among the four emotion categories and reaching 250 samples instead of being limited to only 103. To avoid an over-fitting scenario and wrong recognition rates due to similar samples being wrongly distributed in both training and test sets, the synthetic sequences were placed in the same set of the original sample they were generated from. If an original sample is contained in the test set, during the 10-cross validation approach, its associated forged sequences are not considered for the training set. In Tab.~\ref{augemented-table} the results using the augmented data are reported.

Models trained on the synthetic sequences easily outperform their counterparts trained on the original dataset. These improvements also affect the more challenging frustrated set, clearly demonstrating that having more real samples would improve the classification rates, especially with our proposed network.
\subsection{Comparisons}
The proposed method is compared to the current literature in non-acted body affect recognition. The results on the UCLIC dataset, shown in Tab.~\ref{comparison_Kleinsmith}, highlight how our method exceeds the pioneer work of Kleinsmith et al.~\cite{5704207} in the classification of each class, especially in the recognition of the frustrated affect, that represents one of the major challenges offered by this dataset, the accuracy is greatly increased. Moreover, our solution is in line with respect to the reported human base rate.

In Tab.~\ref{comparison_overall} the overall results based on the average of the accuracy rates, obtained on each class, are shown. One of the competitors does not consider the accuracy related to the frustrated class, this is due to the protocol used in \cite{6250780} that excludes this class because the are few samples. Even if we do not consider these overall values useful to understand the actual performance of a method, we have reported them to perform a comparison with \cite{6250780}. It should be noted that even using this measure our system achieves the best results.
\begin{table}[t]
	\begin{center}
		\resizebox{\columnwidth}{!}{
			\begin{tabular}{l c c c c}
				\hline
				system & C & T & F & D \\
				\hline
				\hline
				human base rate & 57.00\% & 64.00\% & 39.00\% & 61.00\% \\
				Kleinsmith et al. \cite{5704207} & 65.30\% & 61.90\% & 16.00\% & 64.70\% \\
				our & \textbf{78.89\%} & \textbf{68.57\%} & \textbf{35.00\%} & \textbf{68.00\%} \\
				\hline
			\end{tabular}
		}
	\end{center}
	\caption{Experimental comparisons on the UCLIC dataset using the 10-cross validation protocol.}
	\label{comparison_Kleinsmith}
\end{table}
\begin{table}[t]
	\begin{center}	
		\resizebox{0.8\columnwidth}{!}{
			\begin{tabular}{l c c}
				\hline
				system & NoFrustrated  & Overall\\
				\hline
				\hline
				Garber et al. \cite{6250780} & 66.50\% & - \\
				Kleinsmith et al. \cite{5704207}  & 66.33\% & 59.22\% \\
				our LSTM R1+M1 & 68.80\% & 56.60\% \\
				our Full R1+M1, M0  & \textbf{71.82\%} & \textbf{62.61\%} \\
				\hline
			\end{tabular}
		}
	\end{center}
	\caption{Experimental comparisons on the UCLIC dataset with and without the frustrated class based on the overall accuracy.}
	\label{comparison_overall}
\end{table}

\section{Conclusion}
In this paper, an original combination of local and global temporal features is used, with a custom deep neural network architecture, to realize an original non-acted body affect recognition method. An exhaustive experimental phase was performed on the only benchmark dataset available in the current literature. The obtained results shown how the proposed solution outperforms key works in this topic, thus demonstrating how time based features can improve the classification performance with respect to the single postures. As future development, we are investigating new feasible features as well as implementing an alternative solution to generate realistic synthetic data (i.e., by Generative Adversarial Networks (GAN)). To conclude, we are planning to improve the UCLIC dataset by integrating new real samples, especially for the frustrated class, and by defining new affect classes.

{\small
	\bibliographystyle{ieee}
	\bibliography{egbib}
}

\end{document}